\documentclass[11pt]{article}

\usepackage[final]{acl}

\usepackage{times}
\usepackage{latexsym}
\usepackage[T1]{fontenc}
\usepackage[utf8]{inputenc}
\usepackage{microtype}
\usepackage{inconsolata}

\usepackage{graphicx}
\usepackage{booktabs}
\usepackage{multirow}
\usepackage{xcolor}
\usepackage{amsmath}
\usepackage{amssymb}
\usepackage{xspace}
\usepackage{enumitem}
\usepackage{url}

\newcommand{\synca}{SynCA\xspace}
\newcommand{\eg}{e.g.,\xspace}

\newcommand{\fixed}[1]{\textbf{#1}}

\title{Auditing and Repairing LLM-as-Judge Failures\\in a Production Text-to-SQL Pipeline}

\author{
  Haowei Liu \\
  Santa Clara University \\
  \texttt{hliu6@scu.edu} \\\And
  Hsin-Tai Wu \\
  Independent Researcher \\
  \texttt{htwunew@gmail.com} \\\And
  Yi Fang \\
  Santa Clara University \\
  \texttt{yfang@scu.edu} \\
}

\begin{document}
\maketitle

\begin{abstract}
Production text-to-SQL pipelines often end with an LLM-as-judge whose agreement with human annotators has never actually been measured. When we checked ours, the deployed \texttt{gpt-4o-mini} judge agreed with two-author gold at only Cohen's $\kappa = 0.04$ on a disagreement-enriched set and $0.42$ on a uniform-random spot-check, over-flagging 77.1\% of the human-\textsc{faithful} cases in the enriched set. Most of its over-flags trace back to a single mechanism we call \textsc{Grade-Hallucination}. A self-hosted \texttt{Qwen3.6-27B} replacement ($\kappa=0.72$) lands in the same range as \texttt{Claude Opus 4.7} ($\kappa=0.71$); the head-to-head is underpowered at $n=96$, but for the deployment decision that hardly matters, since Qwen costs roughly 1/300 as much per call. Ensembling does not help for free. Pairing the weak judge with a stronger one degrades agreement, whereas three strong judges under unanimity routing reach $\kappa = 0.79$ at 89.7\% auto-coverage. Applied out-of-domain, the same audit recipe flags 25.5\% of BIRD-financial's expert-authored gold SQLs as candidate gold-SQL issues under our annotation protocol. Code and pre-registration are at \url{https://github.com/JamesL404/synca-audit}; the datasets are deferred to the post-publication release for anonymization reasons.
\end{abstract}

\section{Introduction}
\label{sec:intro}

\begin{figure*}[t]
\centering
\includegraphics[width=0.92\linewidth]{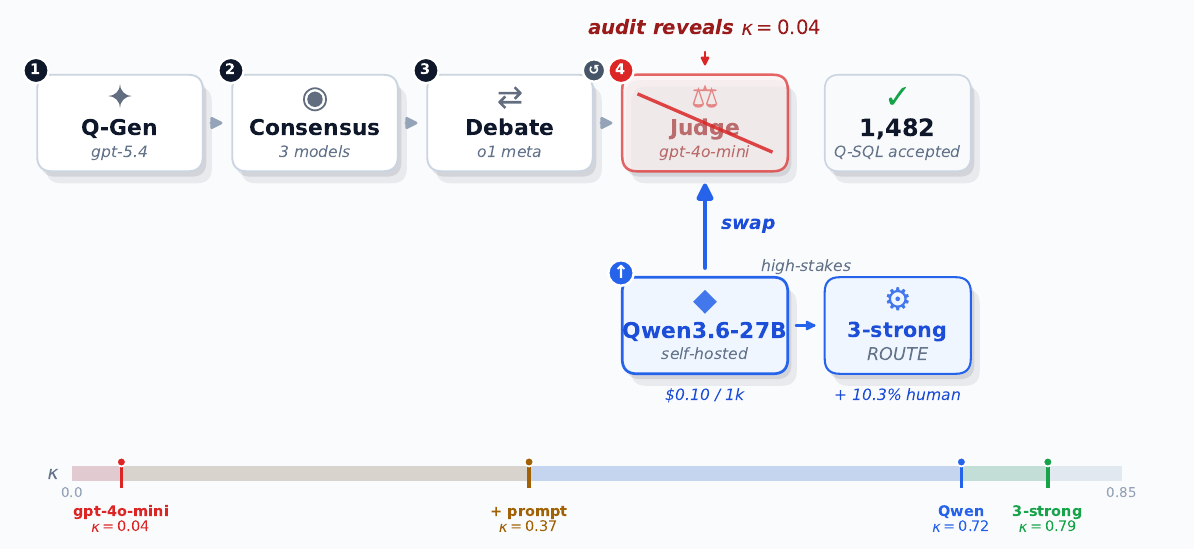}
\caption{The \synca production pipeline (top) and our recommended judge-stage repair (bottom). The production pipeline runs Q-Gen, multi-model consensus, debate, and a final faithfulness judge over 1{,}482 accepted question--SQL pairs. The audit finds the final judge (\texttt{gpt-4o-mini}) at $\kappa=0.04$ versus two-author human gold. The repair replaces it with self-hosted \texttt{Qwen3.6-27B} ($\kappa=0.72$, $\sim$1/300 of Opus's per-call cost); high-stakes deployments route the residual to a 3-strong ensemble reaching $\kappa=0.79$ at 89.7\% auto-coverage.}
\label{fig:pipeline_repair}
\end{figure*}

Industry teams routinely build synthetic benchmarks to evaluate text-to-SQL systems on proprietary databases that no public benchmark covers \citep{spider, bird, spider2}. We design a four-stage pipeline for this (Figure~\ref{fig:pipeline_repair}, top): an LLM generates candidate questions, several LLMs independently produce SQL and their answers are compared, disagreements move into a debate stage, and a final LLM-as-judge scores SQL faithfulness. Every stage filters some errors and every stage is imperfect, and full human review does not scale.

This pipeline, \synca, runs in industrial production over a 604{,}146-row user database with 43 columns. It produced 1{,}482 accepted question--SQL pairs across three difficulty tiers (460 easy / 659 medium / 363 hard). Its output gates an internal analyst-assist tool ($\sim$30 daily users) and a continuous LLM evaluation harness, so a faulty judge propagates to product-visible acceptance rates and to the corpus used to fine-tune in-house models. The repair we report is deployed: self-hosted Qwen now handles routine traffic, 3-strong unanimity routing covers high-stakes runs, and the released artefact is the configuration our internal teams use.

The audit started informally. Domain analysts reviewing the production judge's flagged outputs kept finding plausible queries that the judge had rejected for reasons that fell apart on a closer look. Two authors then annotated 234 questions oversampled from the judge's flagged set. Against this gold, production \texttt{gpt-4o-mini} reaches Cohen's $\kappa = 0.04$; on a uniform-random spot-check of 200 further questions it reaches $\kappa = 0.42$, so the $0.04$ figure is conditional on the enriched sample and is not the typical deployment rate. Its 77.1\% over-flag rate on human-\textsc{faithful} cases in the enriched set concentrates in one mechanism, which we call \textsc{Grade-Hallucination}: the judge sees \texttt{`E'}, \texttt{`z'}, or \texttt{`None'} rows in a result preview, recalls a canonical $\{A,B,C,D\}$ set, and flags the SQL for ``including invalid grades'' that the question never asked it to exclude.

We test three replacements: closed \texttt{gpt-5.4-mini} ($\kappa = 0.63$ [0.49, 0.76]), open \texttt{Qwen3.6-27B} ($\kappa = 0.72$ [0.59, 0.83]), and closed-frontier \texttt{Claude Opus 4.7} ($\kappa = 0.71$ [0.57, 0.83]). A pre-registered active-sampled adjudication on 96 Qwen-vs-Opus disagreements comes out statistically inconclusive (McNemar $p=0.082$; BCa 95\% CI $[-0.198, +0.276]$; a 57:39 lead for Qwen). Pairing the weak production judge with a stronger one degrades agreement ($\kappa = 0.57$ vs $0.63$), while three decorrelated competent judges with unanimity routing reach $\kappa = 0.79$ at 89.7\% auto-coverage.

\paragraph{Contributions.}
(i) We diagnose a production LLM-as-judge that agrees with humans at $\kappa = 0.04$, and show that a self-hosted open-source replacement (\texttt{Qwen3.6-27B}) is not measurably worse than closed-frontier \texttt{Opus 4.7} at our sample size while costing about 1/300 as much per call.
(ii) We isolate the dominant failure mechanism, \textsc{Grade-Hallucination}, and run a four-variant prompt sweep. The mechanism is partly prompt-fixable (the \emph{no-preview} variant raises $\kappa$ from 0.04 to 0.37), but no prompt we tried closes the gap to a stronger judge.
(iii) We show that pairing a weak judge with a stronger one degrades agreement on our data, and report a disagreement-win diagnostic that tells you in advance whether ensembling will help. With two further decorrelated competent judges, unanimity routing reaches $\kappa = 0.79$ at 89.7\% auto-coverage.
(iv) Applied out-of-domain to BIRD-financial, the audit recipe flags 25.5\% of its expert-authored gold SQLs as candidate gold-SQL issues under our annotation protocol; schema enrichment helps opaque-coded schemas (BIRD) but hurts English-named ones (Spider).

\section{Related Work}
\label{sec:related}

\paragraph{LLM-as-judge.} The protocol was popularized by \citet{zheng2023judging}; bias and reliability analyses include \citep{gu2024survey,han2025judges,shi2024position,krumdick2025nofreelabels,wang2023fairevaluators}. \citet{murugadoss2024evaluator} observe that LLM judges impose criteria that are not in the rubric. Our \textsc{Grade-Hallucination} mechanism is a concrete special case of this for SQL faithfulness, with a clear trigger (result-row inspection) and a clean fix (the no-preview prompt variant). Our human-annotation protocol draws on \citep{cheng2024silicon,plank2022humanlabel,dong2024personalized}; we report pre-consensus IAA on the 200-question spot-check ($\kappa=0.87$) and the 96-case active-sampled set ($\kappa=0.934$).

\paragraph{Text-to-SQL evaluation.} Spider \citep{spider, spider2}, BIRD \citep{bird, wretblad2024bird}, and downstream methods \citep{din_sql, dail_sql} have shifted from execution-accuracy toward LLM-based semantic equivalence \citep{taming_sql, recent_t2sql_survey}. We do not propose a new evaluator. Instead we audit a deployed one and document the conditions under which off-the-shelf judges fail. As a side effect, one of our judges catches every BIRD-financial gold bug we identified, which points to the LLM judge as a benchmark-quality auditor in its own right.

\paragraph{Ensembling and cost cascades.} Prior LLM-judge ensembling work \citep{auto_prompt_ensemble, ensemble_disagreement, verga2024poll} reports positive results, mostly on long-form generation where panel diversity captures different judgement axes. Our negative result on weak+strong pairing qualifies that picture: a panel helps only when each member is individually well-calibrated, and otherwise inherits the weak member's noise. Cost-cascade and selective-prediction work \citep{frugal_gpt, yue2023cascade, zellinger2025, soiffer2025, ask_strong_judge, uncertainty_routing, trustorescalate2024} frames the coverage--accuracy Pareto problem; unanimity routing is one operating point on that frontier, here applied to LLM-as-judge rather than to generation quality.

\section{The SynCA Pipeline}
\label{sec:pipeline}

\synca generates and validates synthetic NL$\to$SQL questions over \texttt{users}, a 604{,}146-row SQLite table with 43 columns. The table is a synthetic derivative of an internal telecom customer table. Its schema, value distributions, and data-quality artefacts (text-encoded booleans, sparse-NULL interest columns, non-canonical categoricals) are preserved from production, but every row is independently synthesized. The released table contains no names, phone numbers, account IDs, or join-keys against any non-synthetic table, and row-level synthesis breaks any deterministic mapping back to the original.

The schema deliberately keeps the production-style traps. Booleans are stored as the strings \texttt{`True'}/\texttt{`False'} rather than 0/1. \texttt{data\_usage} is in raw bytes and must be divided by $10^9$ to give GB. Interest columns are over 80\% NULL, where NULL is semantically ambiguous between ``not interested'' and ``unknown.'' Small-cardinality categoricals contain non-canonical values: \texttt{demographics\_income\_trust}, for instance, takes values $\{A, B, C, D, E, z, \texttt{NULL}\}$, where \texttt{E} and \texttt{z} are legitimate codes (``other,'' ``not provided'') in the production schema but trigger \textsc{Grade-Hallucination} in untrained judges.

The pipeline has four stages: (1) Q-Generation via \texttt{gpt-5.4} following self-instruction \citep{self_instruct,evol_instruct}, parameterized by difficulty target (join count, aggregation depth, filter complexity); (2) multi-model consensus among \texttt{gpt-4o}, \texttt{claude-sonnet-4.6}, and \texttt{gemini-2.5-flash}, which produce SQL independently and are accepted when at least two result-row sets match value-only (column-name agnostic); (3) debate, up to two rounds with \texttt{o1} as meta-judge over still-disagreeing cases, classifying outcomes as ambiguous-SQL, ambiguous-question, or unanswerable; and (4) a faithfulness judge (originally \texttt{gpt-4o-mini}) scoring each surviving candidate as \textsc{faithful} / \textsc{minor-issue} / \textsc{unfaithful} given the schema, question, SQL, and a top-10-row executed-result preview. \synca contains 1{,}598 final records, of which 1{,}482 are accepted; the remaining 116 were rejected at consensus or debate (Appendix~\ref{app:pipeline} details the per-stage counts).

Table~\ref{tab:datasplit} lists every \synca subset and the claim each one supports. Three human-gold sets play different roles and are not interchangeable: the disagreement-enriched 234-set carries the calibration measurements, the 200-question uniform-random spot-check supports the population faithful-rate estimate, and the 96-case active sample supports only the Qwen-vs-Opus adjudication. The remaining 1{,}248 accepted records carry automated Qwen verdicts, and 1{,}241 of them additionally carry an Opus verdict. Every $\kappa$ we report names the subset it was computed on, since a number from the enriched set is conditional on the production judge having flagged the question.

\begin{table*}[t]
\centering
\small
\begin{tabular}{llr}
\toprule
Subset & Role / provenance & $n$ \\
\midrule
Final records         & all pipeline outputs                          & 1{,}598 \\
\quad Accepted        & passed consensus+debate (released benchmark)  & 1{,}482 \\
\quad Rejected        & dropped at consensus or debate                & 116 \\
\midrule
234-set               & two-author gold, disagreement-enriched calib. & 234 \\
200 spot-check        & two-author gold, uniform-random ($\kappa{=}0.87$) & 200 \\
382 active sample     & pre-registered stratified draw from the 1{,}248-pool & 382 \\
\quad 96 strict-disagr. & two-author gold, Qwen-vs-Opus disagreements  & 96 \\
\midrule
1{,}248 supplementary & accepted $\setminus$ 234-set; Qwen-judged      & 1{,}248 \\
\quad dual-judged     & also Opus-judged (10 SQL errors excluded)     & 1{,}241 \\
\bottomrule
\end{tabular}
\caption{\synca subsets and their provenance.}
\label{tab:datasplit}
\end{table*}

\section{Human Calibration Set}
\label{sec:calibration}

Throughout, \emph{calibration} means agreement with human labels in the psychometric sense (Cohen's $\kappa$), not probability calibration of a model's confidence scores.

Two authors (PhD researchers with telecom-domain and SQL-evaluation expertise) manually re-evaluated 234 question--SQL pairs drawn primarily from the production judge's flagged set, with a balance of accepted cases; disagreements were resolved by discussion. The disagreement-enriched sampling is intentional, since it concentrates labels where the pipeline could plausibly be contested. We report pre-consensus inter-annotator agreement on two complementary sets: the 200-question uniform-random spot-check (\S\ref{sec:spotcheck}, $\kappa=0.87$) and the 96-case active-sampled adjudication (\S\ref{sec:active_sample}, $\kappa=0.934$). The 234-set distribution is 192 \textsc{faithful} and 42 \textsc{unfaithful}, and 148 of these (63\%) are false-flag corrections. To check reproducibility from outside the project, one screened independent SQL-literate annotator, blind to author labels and model verdicts, re-labeled all 234 cases and agreed with the two-author consensus at $\kappa=0.85$ (95.7\% raw). Author-only annotation remains the load-bearing limitation (\S\ref{sec:limitations}). Appendix~\ref{app:calibration_stats} gives the stratification.

\section{Judge Calibration}
\label{sec:calibration_results}

We evaluate four judges with identical prompts: \texttt{gpt-4o-mini}, \texttt{gpt-5.4-mini}, \texttt{Qwen3.6-27B-Instruct} (served via vLLM \citep{qwen3}), and \texttt{Claude Opus 4.7}. Following \citep{cohen1960, landis1977kappa, han2025judges, murugadoss2024evaluator}, we report Cohen's $\kappa$ as the primary metric and fold \textsc{minor-issue} into \textsc{faithful}. We report $\kappa$ on two cuts of human gold: (a) the 234-set, conditional on production-judge flagging, which is where the failure mode lives; and (b) the 200-question uniform-random spot-check, which is closer to typical deployment. Both are needed, because either cut on its own is misleading.

\begin{table*}[t]
\centering
\small
\begin{tabular}{lcc}
\toprule
Judge & Conditional (n=234) & Unconditional (n=200) \\
\midrule
\texttt{gpt-4o-mini}      & 0.04 [-0.02, 0.09] & 0.42 [0.24, 0.58] \\
\texttt{gpt-5.4-mini}     & 0.63 [0.49, 0.76]  & 0.67 [0.52, 0.80] \\
\texttt{Qwen3.6-27B}      & \fixed{0.72 [0.59, 0.83]} & \fixed{0.94 [0.86, 0.99]} \\
\texttt{Claude Opus 4.7}  & \fixed{0.71 [0.57, 0.83]} & \fixed{0.93 [0.86, 0.99]} \\
\bottomrule
\end{tabular}
\caption{Per-judge $\kappa$ vs two-author consensus gold (bootstrap 95\% CIs, $B=5{,}000$; binary, \textsc{minor-issue} folded to \textsc{faithful}). Qwen$\approx$Opus at this measurement resolution (6--7 split on the 13 conditional disagreements). \texttt{gpt-4o-mini} over-flags (FNR 77.1\%/FPR 14.3\%); \texttt{gpt-5.4-mini} under-flags (FNR 4.7\%/FPR 35.7\%).}
\label{tab:per_judge}
\end{table*}

The production judge is at chance on hard cases ($\kappa=0.04$) but moderate on typical ones ($\kappa=0.42$). Qwen and Opus fall in the same range on both cuts. The head-to-head between them is underpowered (later in this section), so we do not claim parity, only that the cost decision is not driven by any measured calibration gap. A between-judge McNemar on the full $n=382$ active sample (\S\ref{sec:active_sample}) does reject grading-style interchangeability ($b{=}47, c{=}5, p{=}1.3{\times}10^{-9}$), with Qwen preferring strict \textsc{unfaithful} where Opus prefers \textsc{minor-issue}. This still does not resolve the accuracy-vs-human question.

\paragraph{Failure-mode taxonomy.} We pattern-mined the judge's free-text reasons on the 148 false-flag cases. An initial 7-class taxonomy (developed by the two authors without pre-registration) gave second-coder $\kappa=0.29$: the second coder collapsed five minority classes into a single ``phantom-constraint'' bucket, and grade-hallucination was the only subtype with item-by-item agreement. We therefore post-hoc collapsed it to a binary grade-hallucination-vs-other-phantom split ($\kappa = 0.47$ at 68\% second-coder recall on grade-hallucination). This binary split was \emph{not} pre-registered, and the rest of the paper relies only on the binary headline. The 66\% grade-hallucination figure (author labels) survives recoding at 68\% second-coder recall, and both upgraded judges fix over 94\% of these cases. Grade-hallucination is the \synca-specific instance of a broader phantom-constraint family that also surfaces on BIRD-financial (``verification-demanding,'' Appendix~\ref{app:bird_patterns}); see Appendix~\ref{app:failure_modes} for details.

\paragraph{Is the gap prompt-fixable? (H5).} A four-variant prompt sweep on \texttt{gpt-4o-mini} (CoT, explicit-rubric, no-preview, strict; Table~\ref{tab:gpt4o_prompt_ablation}) gives two findings. First, every variant flips the failure mode from over-flagging (0.77) to under-flagging (0.58--0.74): the judge moves off the grade-hallucination attractor, but lands on the opposite failure rather than on calibrated discrimination. Second, the \emph{no-preview} variant is the single most effective change, raising $\kappa$ from 0.04 to 0.37 by removing the offending input rather than asking the model to ignore it. Prompting does substantially reduce the specific mechanism: on the 98 author-labeled grade-hallucination false flags, \emph{no-preview} and \emph{strict} each recover 93/98 and the explicit rubric recovers 87/96 parseable cases. What prompting does not do is close the whole-set calibration gap: the distance to Qwen's $\kappa = 0.72$ remains. We cannot rule out that some unexplored prompt closes it, but four reasonable variants (CoT and an anti-confabulation rubric among them) leave the same residual. Schema/preview and SQL-feature ablations are in Appendices~\ref{app:h5_ablation} and \ref{app:sql_feat}.

\begin{table}[t]
\centering
\footnotesize
\setlength{\tabcolsep}{4pt}
\begin{tabular}{@{}lccc@{}}
\toprule
Variant & $\kappa$ [95\% CI] & over- & under- \\
\midrule
baseline (production) & 0.04 [-0.02, 0.09] & 0.77 & 0.14 \\
CoT                   & 0.32 [\phantom{-}0.16, 0.49] & 0.04 & 0.71 \\
rubric                & 0.33 [\phantom{-}0.16, 0.47] & 0.11 & 0.58 \\
no-preview (best)     & \textbf{0.37 [\phantom{-}0.20, 0.52]} & 0.07 & 0.60 \\
strict                & 0.23 [\phantom{-}0.07, 0.38] & 0.07 & 0.74 \\
\bottomrule
\end{tabular}
\caption{Four-variant prompt sweep on \texttt{gpt-4o-mini} (234-set; BCa 95\% CIs, $B=5{,}000$; binary, \textsc{minor-issue} folded). Every variant flips the failure mode; none closes the gap to Qwen's $\kappa=0.72$.}
\label{tab:gpt4o_prompt_ablation}
\end{table}

\paragraph{Active-sampled head-to-head: Qwen vs Opus.}
\label{sec:active_sample}
To stress-test the parity reading, we pre-registered a stratified active sample of 382 questions from the 1{,}248-pool (stratum $\alpha$: 182 Qwen-flagged; $\beta$: 200 random Qwen-\textsc{faithful}, seed 42; pre-registration details in Appendix~\ref{app:active_sample}). The 96 strict-disagreement cases were independently labeled by two authors (pre-discussion $\kappa=0.934$). Against consensus gold, Qwen matches 57 and Opus 39; McNemar exact $p=0.082$; the BCa 95\% CI on the $\kappa$-difference is $[-0.198, +0.276]$. At $n=96$ the result is statistically inconclusive: rejecting equality at $\alpha=0.05$ with power $0.8$ for a true $\Delta\kappa=0.10$ would need $n \approx 380$. The 57:39 directional lead favors Qwen, at $\sim$1/300 of Opus's cost.

\section{Ensembling Cheap Judges}
\label{sec:ensemble}

The asymmetry above suggests ensembling, but the two-judge ensembles do not work. AND ties \texttt{gpt-5.4-mini} alone ($\kappa=0.572$ vs $0.631$), and OR collapses to the floor ($\kappa=0.067$). On the 154 cases where the two judges disagree, \texttt{gpt-5.4-mini} is correct 142 times against 12. Disagreement here is dominated by weak-judge noise, so adding a noisy judge to a calibrated one strictly hurts. Three competent judges are a different story: on the 19 cases where \texttt{gpt-5.4-mini} and Qwen disagree the split is 12--7, and unanimity routing reaches $\kappa=0.788$ at 89.7\% auto-coverage (Table~\ref{tab:ens}). This gain is an abstention effect rather than a better judge: at full coverage the 3-strong majority rule reaches only $\kappa=0.731$, and routing buys the remaining $0.057$ by declining to decide the 10.3\% of cases where the judges disagree. On the retained subset every strong judge agrees by construction, so the reported $\kappa$ is a selective-prediction figure and is not comparable to a full-coverage number.

\begin{table*}[t]
\centering
\small
\begin{tabular}{lccc}
\toprule
Strategy & $\kappa$ (95\% CI) & UNF F1 & cov. \\
\midrule
\texttt{gpt-5.4-mini} alone        & 0.631 [0.49,0.76] & 0.701 & 1.000 \\
\quad + \texttt{gpt-4o-mini} AND   & 0.572 [0.42,0.71] & 0.640 & 1.000 \\
\quad + \texttt{gpt-4o-mini} OR    & 0.067 [0.02,0.11] & 0.341 & 1.000 \\
\midrule
\texttt{Qwen3.6-27B} alone         & 0.716 [0.59,0.83] & 0.763 & 1.000 \\
\texttt{Claude Opus 4.7} alone     & 0.705 [0.57,0.83] & 0.750 & 1.000 \\
3-strong MAJORITY (drop weak)      & 0.731 [0.60,0.84] & 0.774 & 1.000 \\
Unanimity routing (3-strong)       & \fixed{0.788 [0.64,0.90]} & \fixed{0.815} & \fixed{0.897} \\
Unanimity routing (4-strong)       & 0.833 [0.67,0.96] & 0.889 & 0.290 \\
\bottomrule
\end{tabular}
\caption{Ensemble strategies on the 234 records with verdicts from all four judges. Unanimity routing auto-decides when all $k$ judges agree, sends others to human. The 3-strong rule achieves $\kappa=0.788$ at 89.7\% auto-coverage.}
\label{tab:ens}
\end{table*}

\paragraph{A decision rule.} Before adding a judge to an ensemble, check two things: that its bootstrap $\kappa$ CI on a domain-specific gold set is clearly above zero, and that its disagreements with the other members are split rather than one-sided. A 142--12 disagreement-win split means the weaker judge is simply wrong, so ensembling will degrade performance; a 12--7 split signals genuinely hard cases, where ensembling can help. Our preferred operating point is unanimity routing over $k{\geq}3$ competent judges, giving $\kappa = 0.79$ at 89.7\% auto-coverage, which amounts to a 10\% human-review budget on the cases that matter. A learned router (GBM, 14 features, 5-fold CV) does not beat this by more than Monte-Carlo noise ($\Delta\kappa = +0.002$; Appendix~\ref{app:learned_router}).

\section{Cross-Domain Replication: BIRD and Spider}
\label{sec:replication}

We run all four judges on BIRD-financial \citep{bird} (106 dev questions, opaque codes like \texttt{A3}) and four Spider dev databases \citep{spider, spider2} (164 gold-correct questions, English column names), of which \texttt{concert\_singer} (45 questions) carries the per-judge comparison.

\paragraph{BIRD-financial: auditing expert gold.} BIRD's expert gold is known to have quality issues \citep{wretblad2024bird}. Running our four judges as a benchmark-quality audit puts a concrete number on it. Two authors classified all 65 judge-flagged cases. 27 are genuine bugs in the gold SQL, and many are objective errors verifiable against BIRD's own question and evidence (\eg q110 asks for 1998/9/2 but the gold filters \texttt{date=`1997-08-20'}; q189 returns the youngest where the question asks the oldest). 16 are ambiguous questions where multiple SQLs are defensible, and 22 are over-flags, the cross-domain analog of \textsc{Grade-Hallucination}. We present these as candidate gold issues under our protocol, not a verdict that the experts erred: $23$--$27$ of $106$ ($21.7$--$25.5\%$). An independent SQL-literate annotator, blind to our labels, re-classified all 65 cases at Cohen's $\kappa=0.86$, confirming 23 of the 27 bug labels, matching the independent report above. We hold our own gold to the same bar (IAA in \S\ref{sec:calibration}): no reference SQL, expert-written or not, is clean by assumption.

\paragraph{Per-judge behaviour on BIRD.} Among the four judges, Qwen is Pareto-optimal as a single auditor: it catches 85\% of the 27 identified bugs at only 7.5\% over-flag, whereas the next-best judge (gpt-5.4-mini at 74\% recall) more than doubles the over-flag rate. Rich-schema enrichment helps the most aggressive judge (\texttt{gpt-5.4-mini}: 17.0\%$\to$11.3\%) but barely moves the better-calibrated ones, and it slightly reduces Qwen's bug-detection (85$\to$78); see Appendix~\ref{app:bird_patterns} for the failure-pattern detail.

\begin{table}[t]
\centering
\footnotesize
\setlength{\tabcolsep}{4pt}
\begin{tabular}{@{}lccc@{}}
\toprule
Judge & raw & BUG & OVER (sp.$\to$rch.) \\
\midrule
gpt-4o-mini  & 19.8 & 48 & 5.7\,/\,3.8 \\
gpt-5.4-mini & 46.2 & 74 & 17.0\,/\,11.3 \\
Qwen3.6-27B  & 39.6 & \textbf{85} & 7.5\,/\,8.5 \\
Opus 4.7     & 32.1 & 70 & 7.5\,/\,7.5 \\
\bottomrule
\end{tabular}
\caption{BIRD-financial ($n{=}106$), all \%. \emph{raw}: raw flag rate. \emph{BUG}: bug-detect on 27 buggy gold SQLs (two-author classification of 65 flags). \emph{OVER}: corrected over-flag rate, sparse / rich schema.}
\label{tab:bird}
\end{table}

\paragraph{Spider (4 databases).} Aggregated over four Spider dev DBs ($n=164$ gold-correct questions, English schema), sparse$\to$rich over-flag rates are: gpt-4o-mini 11.0$\to$15.9\%, gpt-5.4-mini 9.8$\to$6.1\%, Opus 1.8$\to$0.6\%. Opus has the lowest or tied-lowest over-flag in every cell across all four DBs. Rich-schema is harmful on English schemas for \texttt{gpt-4o-mini} but helps \texttt{gpt-5.4-mini}; the practical rule is to add rich-schema only when column names are opaque (Appendix~\ref{app:spider}).

\section{Cost and Released Benchmark}
\label{sec:cost}

\paragraph{Cost: two deployment scenarios.} We give per-1{,}000-question rates below (full breakdown in Appendix~\ref{app:cost}; provider list rates, plus an amortized \$0.30/hr A100 80GB for self-hosted Qwen).

\emph{Scenario A --- single-judge replacement (Qwen-only).} Replace the production judge with self-hosted \texttt{Qwen3.6-27B} (\$0.10) instead of \texttt{gpt-4o-mini} (\$0.40); Qwen's \$0.10 is $\sim$1/300 of Opus's \$30. This is the right operating point once domain calibration is established and the goal is a cheap, reliable single gate.

\emph{Scenario B --- 3-strong unanimity routing.} Run gpt-5.4-mini + Qwen + Opus on every question (\$31 / 1k, dominated by the closed APIs) and route the ${\sim}10\%$ non-unanimous cases to humans. Per-pass cost is comparable to Opus alone, but the auto-decided $\kappa$ is 0.79 (versus $\kappa\approx 0.7$ for any single judge). Use this when the precision premium is worth the cost (high-stakes runs, evaluation harnesses, training-corpus curation). Scenario A captures the 300$\times$ saving for routine traffic, and Scenario B trades that saving for higher accuracy.

\paragraph{Final benchmark.} Applying \texttt{Qwen3.6-27B} to the 1{,}248 supplementary records (about 50 min on one GPU via vLLM, 10 unparseable, 0.8\%) yields \synca-final with 1{,}249 \textsc{faithful}, 170 \textsc{unfaithful}, 53 \textsc{minor-issue}, and 10 \textsc{error}, which is 84.3\% \textsc{faithful} as the Qwen-raw operational rate. The 200-question uniform-random spot-check (\S\ref{sec:spotcheck}) stratified-reweights this to a population estimate of about 90.6\% (per-class precisions in Appendix~\ref{app:spotcheck}). We report both: 84.3\% for CI-tracking and 90.6\% for benchmark-quality comparisons (Appendix~\ref{app:reconciliation}).

\paragraph{Spot-check IAA.}
\label{sec:spotcheck}
Two authors independently annotated the 200 spot-check records before any discussion. Pre-consensus Cohen's $\kappa = 0.87$ (96.5\% raw agreement; 7/200 disagreements). Qwen's binary accuracy is 98.0\% (196/200) with $\kappa=0.935$, well above its agreement on the disagreement-enriched 234-set, as expected (full table in Appendix~\ref{app:spotcheck}).

\paragraph{Dual-judge validation on the supplementary 1{,}248 set.}
\label{sec:dual_judge_release}
We also labeled all 1{,}248 supplementary records with Opus under the same prompt ($n=1{,}241$ after excluding 10 SQL errors). Inter-judge raw agreement is 95.5\% (1{,}185/1{,}241), and Cohen's $\kappa = 0.728$ [0.655, 0.791] (BCa, $B=5{,}000$). This sits within the 234-set Qwen-vs-human CI [0.61, 0.84], so the calibration audit transfers to the supplementary set. Of the 56 discordant cases, 44 are Qwen-flagged and Opus-\textsc{faithful} (mild borderline conservativeness, \S\ref{sec:spotcheck}).

\paragraph{Release.} The audit-pipeline code, pre-registration, prompt templates, decoding settings, and analysis scripts are available now at \url{https://github.com/JamesL404/synca-audit}. The datasets (the synthetic \texttt{users.sqlite} database; the 1{,}482-record \texttt{benchmark\_v2\_final\_judged.json} with per-record provenance; the 234-set, 200-spotcheck, and 96-active-sample human-gold files; the dual-judged supplementary 1{,}248; the BIRD and Spider cross-domain judged files; and \texttt{MODELS.md} with exact snapshot identifiers) are deferred to the post-publication release for anonymization reasons.

\section{Discussion}
\label{sec:discussion}

\synca's judge stage gates a SQL-generation pipeline feeding an internal analyst-assist tool ($\sim$30 daily users) and a continuous LLM evaluation harness; a faulty judge propagates to product-visible acceptance rates and to the corpus used to fine-tune in-house models. The audit began with analyst complaints rather than a planned reliability study. Since then we have replaced the production judge with self-hosted Qwen for routine traffic and added 3-strong unanimity routing for high-stakes runs; the released artefact is the configuration our internal teams now use. We re-audit each judge against a fresh domain-specific gold set whenever the schema or a model snapshot changes, rather than on a fixed calendar. The $n=96$ adjudication remains inconclusive (a 57:39 directional lead for Qwen), but it does not change the deployment decision, since cost dominates.

The audit recipe transfers across domains; the per-judge over-flag rates do not. A judge calibrated on one schema can over-flag on another, depending on column-name opacity. A fresh disagreement-win check on a small gold set is the cheap first move before scaling to a new schema.

\section{Conclusion}
\label{sec:conclusion}

Our production \texttt{gpt-4o-mini} judge agreed with humans at $\kappa = 0.04$, almost entirely because of grade hallucination. Self-hosted \texttt{Qwen3.6-27B} reaches $\kappa = 0.72$ at $\sim$1/300 of \texttt{Claude Opus 4.7}'s per-call cost; the $n=96$ head-to-head is underpowered, but the deployment decision turns on cost. Unanimity routing over three strong judges pushes auto-decided $\kappa$ to 0.79 at 89.7\% coverage, while prompt engineering tops out at $\kappa = 0.37$. Two takeaways generalise: a deployed LLM-as-judge is not safe to trust without periodic calibration against domain-specific human gold, and ensembling cheap judges pays off only when each member is well-calibrated.

\clearpage
\section*{Limitations}
\label{sec:limitations}

\paragraph{Scope and reproducibility.} \synca is a single English-language telecom deployment over SQLite. Cross-domain replication on BIRD-financial and four Spider DBs helps, but per-judge over-flag rates do shift across domains (\S\ref{sec:replication}), so which judge is ``best'' is conditional on the audited deployment; Qwen on Spider is also restricted to \texttt{concert\_singer} due to endpoint outages. OpenAI snapshots are pinned (Appendix~\ref{app:models}); Anthropic and Google judges do not expose dated identifiers, so the closed-judge $\kappa$ values may drift without notice, even though the audit recipe itself is reproducible.

\paragraph{Author-only annotation.} The headline gold was produced by the two paper authors (pre-consensus $\kappa=0.87$ on the 200-spot-check, $\kappa=0.934$ on the 96-active-sample). The independent re-labeling reported in \S\ref{sec:calibration} ($\kappa=0.85$, 95\% BCa CI $[0.75, 0.93]$, $B{=}5{,}000$; 10 disagreements skewed \textsc{unfaithful}, 7/10) shows the gold is not idiosyncratic to the author pair, though over-flag and bug-detect rates are still co-defined with them. We deliberately did not fine-tune any judge on the 234-set, since training and evaluation labels would then share annotators.

\bibliography{refs}

\appendix

\section{Extended discussion}
\label{app:extended_discussion}

\paragraph{What does ``calibration'' mean here?} We use ``calibration'' in its psychometric sense, agreement with humans on a domain-specific gold set, rather than the temperature-or-confidence sense common in the classifier-output literature. No judge weights are tuned. The calibration we measure is a property of (judge, prompt, domain) jointly, not of the judge alone. \texttt{gpt-4o-mini}'s $\kappa=0.04$ is about the production-deployed judge-prompt-domain combination; the same model with a different prompt reaches $\kappa=0.37$ (\S\ref{sec:calibration_results}). Re-auditing on each new domain or prompt is the unavoidable cost.

\paragraph{Why three judges and not five?} Unanimity routing stops paying off quickly. Adding the fourth judge (\texttt{gpt-4o-mini}) drops auto-coverage from 89.7\% to 29\%, because the weak judge frequently disagrees with the three strong ones without changing the auto-decided $\kappa$ enough to justify the extra human-review cost. What helps is decorrelated competent judges; adding more weak ones does not.

\paragraph{Cross-domain transfer (per-judge).} \texttt{gpt-5.4-mini} 4.7\%$\to$17.0\% (3.6$\times$); Qwen 3.7\%$\to$7.5\% (2$\times$); Opus 1.6\%$\to$7.5\% (4.7$\times$). The same judge is calibrated, mid-calibrated, or over-aggressive depending on schema and value-vocabulary.

\section{Model versions and access details}
\label{app:models}

All experiments used the following model identifiers and access modes (access dates 2026-04 through 2026-05). Where the provider exposes a dated snapshot identifier we pin it; OpenAI does, Anthropic and Google do not.
\begin{itemize}[leftmargin=*,topsep=2pt,itemsep=1pt]
  \item \texttt{gpt-4o-mini-2024-07-18} --- OpenAI API. The production audit predates the experiment window; for re-runs we use this snapshot.
  \item \texttt{gpt-5.4-mini-2026-03-17} --- OpenAI API.
  \item \texttt{Claude Opus 4.7} --- Anthropic API endpoint \texttt{claude-opus-4-7}; Anthropic does not currently expose a dated snapshot identifier through the API. Anthropic Python SDK \texttt{anthropic>=0.39}.
  \item \texttt{Qwen3.6-27B-Instruct} --- HuggingFace weights at the public release revision, served via vLLM 0.6.x on a single A100 80GB. \texttt{enable\_thinking=False}; \texttt{max\_tokens=4096}; we strip any \texttt{<think>\ldots</think>} reasoning preamble from outputs. Tensor-parallel size 1, KV-cache enabled, no quantization. OpenAI-compatible HTTP endpoint.
  \item \texttt{gpt-4o-2024-11-20} (consensus generator), \texttt{claude-sonnet-4.6} (no dated snapshot), \texttt{gemini-2.5-flash} (no dated snapshot) --- used only inside the upstream \synca consensus stage (\S\ref{sec:pipeline}); not measured directly.
  \item \texttt{o1-2024-12-17} --- OpenAI API, used only in the \synca debate meta-judge.
\end{itemize}
Decoding: temperature $0$, \texttt{max\_tokens} 1024 for all judges (4096 for Qwen owing to reasoning-mode artefacts). Identical message structures across providers modulo each vendor's role-message conventions. Closed-API judges without a dated snapshot (Anthropic, Google) are subject to silent model drift; the reported $\kappa$ values are point measurements on the access dates above. Prompt templates, decoding settings, and the Qwen HuggingFace revision SHA are in the released artefact (see §\ref{sec:cost}).

\section{Pipeline statistics}
\label{app:pipeline}

\synca contains 1{,}598 records (1{,}482 accepted, 116 rejected). Accepted-difficulty split: 460 easy / 659 medium / 363 hard. Consensus uses \texttt{gpt-4o}, \texttt{claude-sonnet-4.6}, \texttt{gemini-2.5-flash} with value-only row matching; debate uses up to 2 rounds with the \texttt{o1} meta-judge. After all stages, 885 reached 3/3 consensus and 604 reached 2/3.

\section{Cost breakdown and benchmark distribution}
\label{app:cost}

\paragraph{Token accounting.} Per-question token counts measured on a uniformly sampled 50-question subset of \synca: prompt mean $2{,}047$ tokens (p50 $1{,}998$; p95 $2{,}412$); completion mean $284$ tokens (p50 $271$; p95 $441$). Decoding settings: temperature$=0$ for all judges, max\_tokens$=1024$, no top-k or nucleus modification. The prompt is identical across judges modulo each provider's role-message convention; the schema description is fixed at $\sim$1{,}120 tokens, and the question + SQL + result preview occupies $\sim$870 tokens on average.

\paragraph{Hardware and throughput (Qwen).} Self-hosted \texttt{Qwen3.6-27B} runs under vLLM 0.6.x on a single NVIDIA A100 80GB (tensor-parallel size 1, attention KV-cache enabled, no quantization), serving requests through an OpenAI-compatible HTTP endpoint at $\sim$200 questions/min ($\sim$3.3 qps) under serial workload and $\sim$520 questions/min ($\sim$8.7 qps) under 4-way concurrency. GPU utilization sits at 70--85\% during throughput runs. Amortized GPU cost is computed at \$0.30/GPU-hr (US cloud spot-instance rate; varies $\pm$30\% by provider/region).

\paragraph{Cloud rates.} As of submission date: \texttt{gpt-4o-mini} \$0.15/M input + \$0.60/M output; \texttt{gpt-5.4-mini} \$0.50/M input + \$1.20/M output (approximated, vendor-specific); \texttt{Claude Opus 4.7} \$15/M input + \$75/M output. We apply these to our measured token usage to derive per-1{,}000-question costs.

\paragraph{Per-1{,}000-question costs.} \texttt{Qwen3.6-27B} \$0.10 (1$\times$); \texttt{gpt-4o-mini} \$0.40 (4$\times$); \texttt{gpt-5.4-mini} \$1.30 (13$\times$); \texttt{Claude Opus 4.7} \$30 (300$\times$); 3-strong MAJORITY (all three on every question) \$31 (310$\times$); unanimity routing (3-strong on every question; human reviewer on $\sim$9.5\%) \$31 + \$15--30 of human labor (at \$10--20/hr and $\sim$3 questions/min). Per-question prompt/completion token assumptions affect costs roughly linearly; doubling completion length increases Opus cost by $\sim$40\% and gpt-4o-mini by $\sim$30\%.

\paragraph{Sensitivity.} If Qwen's amortized GPU cost is \$1.50/GPU-hr (\eg on-demand H100), the per-1{,}000-question cost rises to \$0.50, still $\sim$60$\times$ cheaper than Opus. The cost-side conclusion does not depend on the specific GPU rate assumption; it depends on Opus's $\sim$\$30/1k API price, which is the dominant term.

\paragraph{Benchmark distribution.} \synca-final verdict distribution on the 1{,}482 accepted records: \textsc{Faithful} 1{,}249 (84.3\%); \textsc{Unfaithful} 170 (11.5\%); \textsc{Minor-Issue} 53 (3.6\%); \textsc{Error} 10 (0.7\%). Provenance: 1{,}248 Qwen-judged + 234 two-author human gold. The supplementary 1{,}248 additionally carry a dual Qwen+Opus verdict pair (the Opus run completed in 6{,}071 s with 0 ERRORs at 4.85 s/call serial).

\section{Calibration set stratification}
\label{app:calibration_stats}

The 234-set spans 52/115/67 easy/medium/hard and 124/110 2-of-3 vs 3-of-3-consensus cases. 148 (63\%) are false-flag corrections (production marked \textsc{unfaithful}; consensus is \textsc{faithful}).

\section{Failure-mode taxonomy: full categories and regex examples}
\label{app:failure_modes}

Pattern-mining the \texttt{judge\_issues} text on the 148 false-flag cases via keyword-pattern categorization yields six categories covering 78\% of cases: grade hallucination (66\%, named in the main text), phantom-constraint (58\%, overlapping), null-handling (27\%), missing-LIMIT (14\%), preview-quirk (9\%), and percentage-denominator (5\%). Regex anchors include patterns like \texttt{(invalid|unexpected) (grade|value)}, \texttt{should (use|filter|include).*?(NULL|IS NULL)}, and \texttt{should (use|add) (LIMIT|TOP)}. Both upgraded judges fix more than 94\% of grade-hallucination cases.

\paragraph{Second-coder reliability.} We re-labeled all 148 false-flag cases using \texttt{gpt-4o-mini} as an independent second coder (a different model from any judge under audit, same 6-category options, deterministic decoding). Cohen's $\kappa$ against the authors' regex labels: $0.291$ [0.194, 0.385] on the 7-class assignment (Landis-Koch ``fair''), and $0.471$ [0.333, 0.606] on the binary grade-hallucination-versus-other split (``moderate''). The disagreement is structural rather than random: the second coder collapses the four minority categories (null-handling, missing-LIMIT, preview-quirk, percentage-denominator; 28 cases total) into a single phantom-constraint category, which suggests they are better understood as named variants of one phantom-constraint mechanism than as cleanly separable categories. The headline grade-hallucination statistic (66\% of over-flags) survives this re-coding at second-coder recall $68\%$. We treat the four sub-categories as illustrative qualitative groupings rather than as a robust taxonomy.

\section{H5 schema/preview ablation on gpt-5.4-mini}
\label{app:h5_ablation}

Removing schema from \texttt{gpt-5.4-mini}'s prompt drops $\kappa$ by 0.20; removing only the executed-result preview drops it by $\sim$0.05. The preview adds little calibration signal, but it is what triggers grade hallucination on the weaker judge.

\section{Prompt-ablation: variant prompts}
\label{app:prompt_variants}

The four prompt-ablation variants tested on \texttt{gpt-4o-mini} (results in Table~\ref{tab:gpt4o_prompt_ablation} of the main text). \emph{CoT} adds ``Think step-by-step before producing the JSON verdict''. \emph{Rubric} adds an explicit ``UNFAITHFUL means\ldots; NOT UNFAITHFUL means\ldots'' rubric forbidding confabulation. \emph{No-preview} removes the result-row preview, isolating the grade-hallucination trigger. \emph{Strict} adds ``Be conservative; flag only if you can point to a specific question-stated constraint that the SQL violates.'' Every variant flips the failure mode from over- to under-flagging; none closes the gap to Qwen's $\kappa=0.72$.

\section{SQL-feature stratification}
\label{app:sql_feat}

Stratifying by SQL features shows \texttt{CASE WHEN} as the dominant residual failure mode (Qwen $\Delta\kappa=-0.32$); \texttt{HAVING} and subqueries do not hurt.

\section{Active-sampled adjudication: pre-registration details}
\label{app:active_sample}

Pre-registered stratified active sample of 382 questions: stratum $\alpha$ = all 182 Qwen-flagged; $\beta$ = 200 random Qwen-\textsc{faithful} (seed 42); pool hash \texttt{a7dee6ea713f9531}. Pre-registered analysis: McNemar's exact test and a BCa bootstrap 95\% CI on $\kappa_{\text{Qwen}} - \kappa_{\text{Opus}}$ ($B=5{,}000$ paired). 96 strict-disagreement cases independently labeled by two authors (3/96 pre-discussion disagreements, $\kappa=0.934$). Bidirectional CI reading: the $[-0.198, +0.276]$ interval excludes neither direction beyond an absolute $\Delta\kappa=0.20$ effect, so a meaningful per-judge quality gap in either direction is not ruled out at this $n$.

\section{Learned routing as a negative control}
\label{app:learned_router}

A GBM trained on 14 features (per-judge verdicts, reasoning-length, SQL and question structure, difficulty) via 5-fold CV on the 234-set reaches coverage 91.0\% at $\kappa=0.790$, an improvement of $\Delta\kappa=+0.002$ and $\Delta$cov$=+1.3$pp over the unanimity rule (89.7\% / 0.788). This sits within Monte-Carlo noise. ROC-AUC against route-required is 0.716, but only 18 positives exist, and the top features are judge-implementation-specific reasoning-length signals unlikely to transfer. On the 96-set the learned router collapses to negative $\kappa$ at every threshold. The reading is that unanimity routing is close to the empirical Pareto frontier of this hypothesis class.

\section{BIRD failure-mode patterns and recovered-vs-not-recovered}
\label{app:bird_patterns}

On BIRD-financial we see a recurring family of phantom-constraint mistakes that do not match grade hallucination exactly but rhyme with it. The most common pattern is what we informally call \emph{verification-demanding}: the judge insists that the SQL prove a literal value (a year, a category code) actually exists in the database, and flags any SQL that omits a defensive existence check. The others (over-eager \texttt{DISTINCT}, over-strict reading of natural-language quantifiers, percentage-scope confusion, NULL handling) are less frequent and largely overlap. Of the 7 \texttt{gpt-5.4-mini} cases that rich-schema enrichment recovers, all 7 are verification-demanding; the 11 that stay flagged are interpretive demands the schema cannot address. \texttt{gpt-5.4-mini}'s corrected over-flag rate (17.0\%) is $\sim$3.6$\times$ its \synca-conditional rate, so judge agreement is domain-specific.

\section{Spider concert\_singer per-judge results}
\label{app:spider}

Table~\ref{tab:spider} gives per-judge false-positive rates (\textsc{unfaithful}-only) across four Spider dev databases under sparse and rich schema. Qwen was available only for \texttt{concert\_singer} (\texttt{gpu7} was unreachable for the other three), where its sparse over-flag is $\sim$0, replicating the in-domain \synca finding. Opus has the lowest or tied-lowest over-flag rate in every (DB, schema) cell, with an aggregate at or below 2\% in both schemas.

\begin{table}[h]
\centering
\footnotesize
\setlength{\tabcolsep}{4pt}
\begin{tabular}{@{}llccc@{}}
\toprule
DB ($n$) & judge & FP$_s$ & FP$_r$ \\
\midrule
concert\_singer (44) & 4o-mini  & 11.4 & 22.7 \\
                     & 5.4-mini & 9.1  & 6.8  \\
                     & Qwen     & \textbf{0.0} & 9.1 \\
                     & Opus     & \textbf{0.0} & \textbf{0.0} \\
\midrule
car\_1 (40)          & 4o-mini  & 17.5 & 12.5 \\
                     & 5.4-mini & 22.5 & 10.0 \\
                     & Opus     & 2.5  & \textbf{0.0} \\
\midrule
pets\_1 (40)         & 4o-mini  & 10.0 & 22.5 \\
                     & 5.4-mini & 7.5  & 7.5  \\
                     & Opus     & 5.0  & 2.5  \\
\midrule
world\_1 (40)        & 4o-mini  & 5.0  & 5.0  \\
                     & 5.4-mini & \textbf{0.0} & \textbf{0.0} \\
                     & Opus     & \textbf{0.0} & \textbf{0.0} \\
\midrule
\textbf{Aggregate ($n=164$)} & 4o-mini  & 11.0 & 15.9 \\
                     & 5.4-mini & 9.8  & 6.1  \\
                     & Opus     & \textbf{1.8} & \textbf{0.6} \\
\bottomrule
\end{tabular}
\caption{Spider 4-DB false-positive rates (\%, \textsc{unfaithful}-only), sparse / rich schema. \texttt{concert\_singer} additionally classifies q16 (the one buggy gold).}
\label{tab:spider}
\end{table}

Qwen's 4 rich-schema FPs on concert\_singer are defensible strict-interpretations (\eg a real \texttt{TEXT}/\texttt{INT} mismatch in \texttt{Stadium\_ID}), not grade-hallucination cases.

\section{Reconciliation: 84.3\% vs 90.6\% in full}
\label{app:reconciliation}

The Qwen-verdict distribution on the 1{,}241-pool is 85.3\% \textsc{faithful} / 10.4\% \textsc{unfaithful} / 4.3\% \textsc{minor-issue}. Stratified-reweighting the per-class precision from Table~\ref{tab:spotcheck} (100\% \textsc{faithful}; 100\% \textsc{minor-issue} folded to \textsc{faithful}; 90\% \textsc{unfaithful}) yields $0.853\times1.0 + 0.043\times1.0 + 0.104\times0.10 \approx 0.906$. Equivalently, $84.3 + 3.6 + 1.2 \approx 90.6$.

\section{200-question spot-check per-class results}
\label{app:spotcheck}

Table~\ref{tab:spotcheck} reports Qwen3.6-27B's per-class precision against two-author gold on the 200-question uniform-random spot-check, with Wilson score 95\% CIs. All four disagreements are Qwen-\textsc{unfaithful} on human-\textsc{faithful} cases, so Qwen is mildly conservative on the borderline. The two 100\% cells rest on small $n$: the \textsc{minor-issue} interval reaches down to 83.9\%, so these are not exact per-class accuracies.

\begin{table}[h]
\centering
\footnotesize
\setlength{\tabcolsep}{4pt}
\begin{tabular}{lrrr}
\toprule
Qwen verdict & $n$ & match & precision [95\% CI] \\
\midrule
\texttt{FAITHFUL}     & 140 & 140 & \fixed{100.0\% [97.3, 100]} \\
\texttt{MINOR\_ISSUE} & 20  & 20  & \fixed{100.0\% [83.9, 100]}\textsuperscript{$\dagger$} \\
\texttt{UNFAITHFUL}   & 40  & 36  & \fixed{90.0\% [76.9, 96.0]} \\
\bottomrule
\end{tabular}
\caption{Qwen3.6-27B per-class precision on the 200-question spot-check (Wilson 95\% CIs). \textsuperscript{$\dagger$}\textsc{minor-issue} folded to \textsc{faithful}.}
\label{tab:spotcheck}
\end{table}

\section{Opus prompt-variation robustness check}
\label{app:opus_prompt_robustness}

Mirroring the Qwen prompt-variation check (Appendix~\ref{app:prompt_robustness}), we tested whether \texttt{Claude Opus 4.7}'s $\kappa$ on the 234-set is stable across prompt variants. Four variants (CoT, rubric, no-preview, strict) were run against the same human gold under deterministic decoding ($T{=}0$, max\_tokens$=1024$, identical message structure). Results (BCa 95\% CIs, $B=2{,}000$): baseline $\kappa=0.706$ [0.559, 0.818]; CoT $\kappa=0.790$ [0.673, 0.886]; rubric $\kappa=0.686$ [0.543, 0.804]; no-preview $\kappa=0.638$ [0.484, 0.763]; strict $\kappa=0.731$ [0.590, 0.836]. All CIs heavily overlap, so Opus is prompt-stable on this set, consistent with the Qwen prompt-robustness finding (\S\ref{app:prompt_robustness}).

\section{Qwen prompt-variation robustness check}
\label{app:prompt_robustness}

\begin{table}[h]
\centering
\small
\begin{tabular}{lccc}
\toprule
Variant & n & $\kappa$ vs gold & $\kappa$ vs baseline \\
\midrule
baseline      & 234 & 0.716 [0.59, 0.83] & 1.000 \\
\texttt{cot}  & 234 & \textbf{0.789 [0.67, 0.89]} & 0.807 \\
\texttt{rubric} & 234 & 0.754 [0.63, 0.86] & 0.871 \\
\texttt{no\_preview} & 234 & 0.673 [0.52, 0.80] & 0.828 \\
\bottomrule
\end{tabular}
\caption{Qwen3.6-27B prompt-variation $\kappa$ on the 234-set (BCa CIs, $B=5{,}000$). All four CIs overlap; Qwen verdicts are stable across prompt choice.}
\end{table}

\section{Optimization trajectory}
\label{app:traj}

\begin{figure}[t]
\centering
\includegraphics[width=\linewidth]{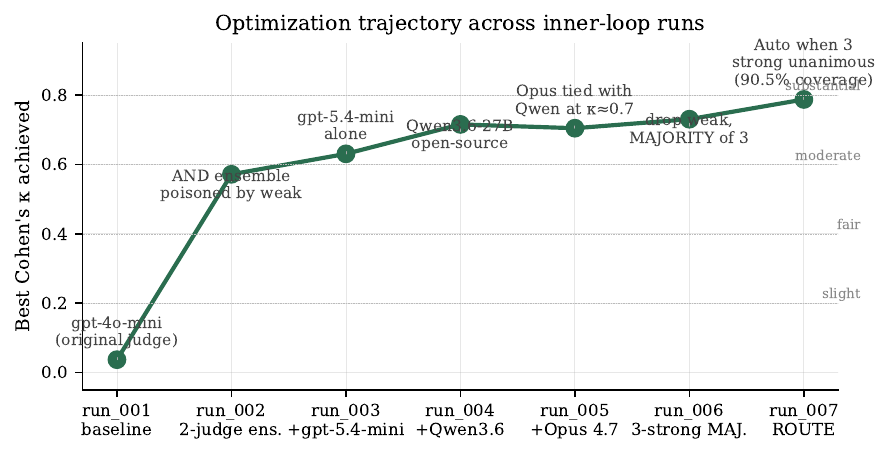}
\caption{Optimization trajectory: best $\kappa$ across the seven main runs.}
\label{fig:traj}
\end{figure}

Figure~\ref{fig:traj} traces the best $\kappa$ across the seven inner-loop runs. The two qualitatively important transitions are the 2-judge regression ($\kappa=0.57$ at run 2) and the routing strategy enabled by the third decorrelated judge.

\begin{figure}[t]
\centering
\includegraphics[width=\linewidth]{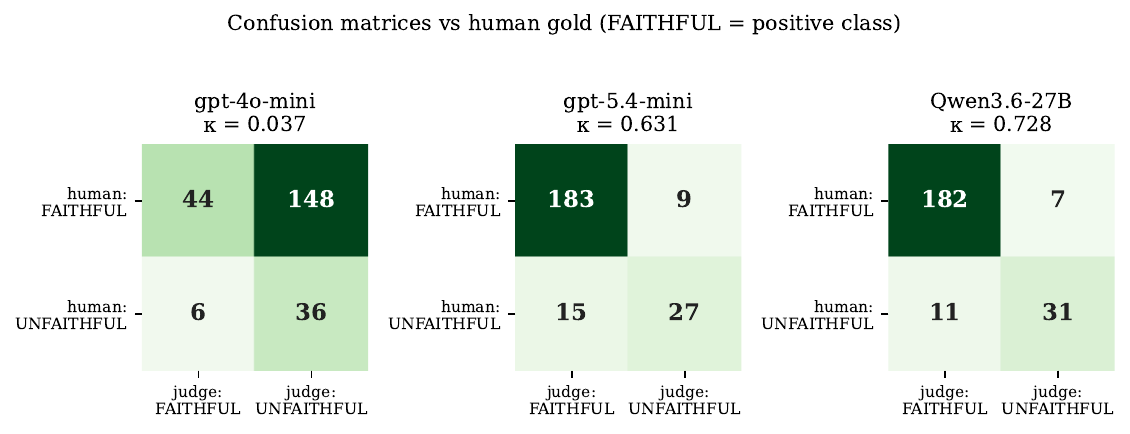}
\caption{Confusion matrices vs human gold (\textsc{faithful} = positive). \texttt{gpt-4o-mini} produces 148 false flags; Qwen has the fewest errors of both types.}
\label{fig:confusion}
\end{figure}

\begin{figure}[t]
\centering
\includegraphics[width=\linewidth]{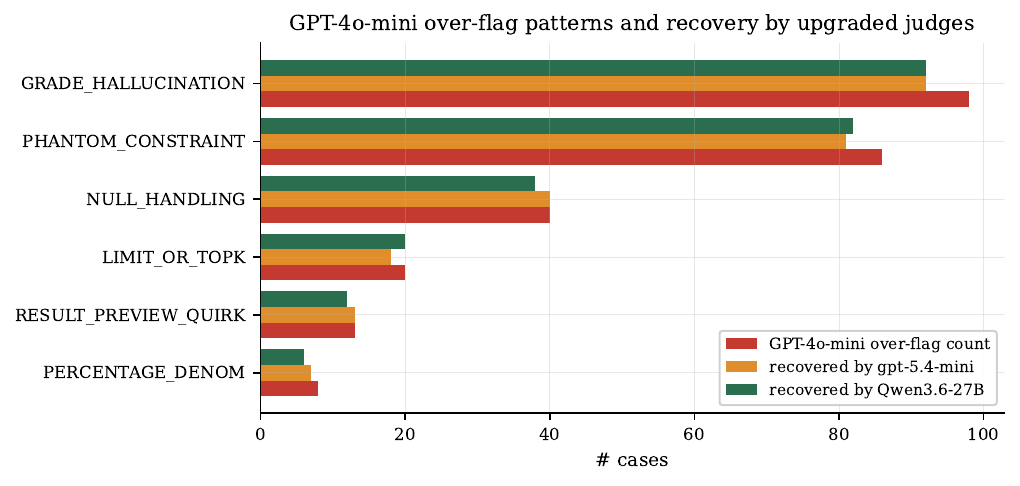}
\caption{Over-flag failure-mode taxonomy. Grade hallucination dominates (66\%); both upgraded judges fix nearly every category.}
\label{fig:failure}
\end{figure}

\begin{figure}[t]
\centering
\includegraphics[width=\linewidth]{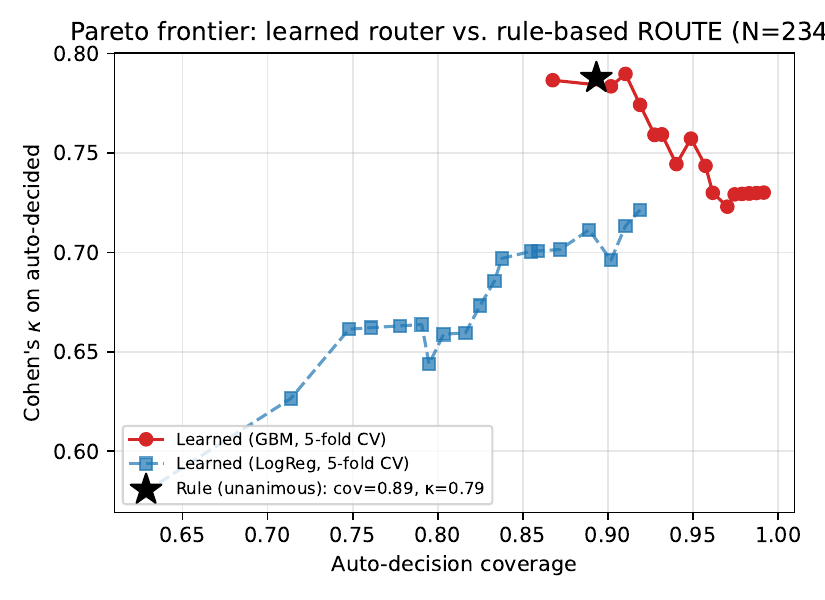}
\caption{Coverage--$\kappa$ Pareto frontier: learned router (5-fold CV on 234-set) versus unanimity routing. The learned router's best point (91.0\% / 0.790) is within Monte-Carlo noise of the rule's (89.7\% / 0.788).}
\label{fig:learned_router_pareto}
\end{figure}

\end{document}